\documentclass[10pt,twocolumn]{article}

\usepackage[margin=0.75in]{geometry}
\usepackage[T1]{fontenc}
\usepackage[utf8]{inputenc}
\usepackage{times}
\usepackage{microtype}
\usepackage{amsmath,amssymb,amsfonts,mathtools}
\usepackage{booktabs,multirow,array}
\usepackage{graphicx}
\usepackage{tikz}
\usetikzlibrary{positioning, fit, calc}
\usepackage{xcolor}
\usepackage{hyperref}
\usepackage{enumitem}

\hypersetup{
  colorlinks=true,
  linkcolor=blue!50!black,
  citecolor=blue!50!black,
  urlcolor=blue!60!black
}

\newcommand{\AFBO}{\textsc{AFBO}}
\newcommand{\OQC}{\textsc{OQC}}
\newcommand{\OQCg}{\textsc{OQC}-gated}
\newcommand{\OQCLR}{\textsc{OQC-LR}}
\newcommand{\PR}{\textsc{PR}}

\title{Orthogonal Quadratic Complements for Vision Transformer Feed-Forward Networks}
\author{Wang Zixian\\
China Mobile Communications Group Shandong Co., Ltd.\\
{\tt\small wangzixian@sd.chinamobile.com}
}
\date{}

\begin{document}
\maketitle

\begin{abstract}
Recent bilinear feed-forward replacements for vision transformers can substantially improve accuracy, but they often conflate two effects: stronger second-order interactions and increased redundancy relative to the main branch. We study a complementary design principle: auxiliary quadratic features should contribute only information that is not already captured by the dominant hidden representation. To this end we propose \emph{Orthogonal Quadratic Complements} (\OQC), which build a low-rank quadratic auxiliary branch and explicitly project it onto the orthogonal complement of the main branch before injection. We also study an efficient low-rank realization (\OQCLR) and gated extensions (\OQC-static and \OQC-dynamic). Under a parameter-matched Deep-ViT/CIFAR-100 protocol with a fixed penultimate-residual readout, full \OQC\ improves an \AFBO\ baseline from $64.25 \pm 0.22$ to $65.59 \pm 0.22$, while \OQCLR\ reaches $65.52 \pm 0.25$ with a substantially better speed--accuracy trade-off. On TinyImageNet, the gated extension \OQC-dynamic\ achieves $51.88 \pm 0.32$, improving the baseline ($50.45 \pm 0.21$) by $+1.43$ points and outperforming all ungated variants. Mechanism analyses show near-zero post-projection auxiliary--main overlap together with improved representation geometry and class separation. The full family---ungated and gated alike---generalizes consistently across both datasets.
\end{abstract}

\section{Introduction}
Feed-forward networks (FFNs) remain one of the main bottlenecks in vision transformers. While simple MLP blocks carry most of the channel mixing capacity, they struggle to capture higher-order interactions explicitly. Consequently, ongoing research often replaces or augments standard FFNs with more expressive multiplicative or bilinear operators. However, we argue that the true bottleneck is not solely the lack of capacity, but rather the failure to extract structurally novel information. 

Simply adding an auxiliary high-order branch to a host FFN does not automatically guarantee better information utilization. Whether augmenting a standard MLP or a state-of-the-art bilinear operator (such as \AFBO), a recurring failure mode is representation redundancy: additional quadratic branches often collapse back into the dominant main representation. This suggests that the key problem is not merely ``add another second-order branch'', but rather \emph{how to inject non-redundant complementary information}.

We approach this from a finite-dimensional Hilbert-space viewpoint. Let the main hidden map define the dominant subspace of a transformer block. An auxiliary quadratic branch should then act as a complement, not as a duplicate. This motivates \OQC: we first construct a low-rank quadratic feature, then explicitly remove the component aligned with the main branch, and finally inject only the orthogonal complement. The resulting module can be instantiated as a full hidden-space complement (\OQC) or as a cheaper low-rank variant (\OQCLR). We also explore gated versions (\OQC-static and \OQC-dynamic), which modulate the complement after orthogonalization.

Across the experiments in this paper, the picture is consistent. Every complement variant improves the \AFBO\ baseline on both CIFAR-100 and TinyImageNet. Full \OQC\ provides the most geometry-preserving improvement, \OQCLR\ is the best efficiency-oriented approximation, and the \OQCg\ family generalizes reliably across both datasets---with \OQC-dynamic\ achieving the highest TinyImageNet accuracy. Crucially, \OQC\ serves as a broad enhancement pattern: our generic transfer study shows that orthogonal quadratic complements provide substantial performance boosts to standard MLPs, while continuing to yield reliable gains even when applied to highly-optimized bilinear operators like \AFBO.

\paragraph{Contributions.}
\begin{itemize}[leftmargin=*,itemsep=2pt,topsep=2pt]
\item We introduce \OQC, an FFN design principle that injects only the orthogonal quadratic complement of the dominant main branch instead of an unrestricted auxiliary quadratic branch.
\item We derive two practical realizations: a full complement model with the best cross-dataset accuracy, and an efficient low-rank variant (\OQCLR) with a better speed--accuracy trade-off.
\item We provide mechanism evidence that the complement projection removes auxiliary--main overlap while improving representation geometry and class separation, and we delineate where the method does and does not transfer beyond the primary \AFBO\ host backbone.
\item We additionally study gated extensions (\OQC-static and \OQC-dynamic) and show that adaptive gating is promising on CIFAR-100 but currently does not displace full \OQC\ as the strongest cross-dataset claim.
\end{itemize}

\section{Related Work}
\paragraph{Vision transformers.}
The Transformer architecture~\cite{vaswani2017attention} was originally proposed for sequence modelling and has since become a standard backbone for visual recognition via ViT~\cite{dosovitskiy2021vit}. Follow-up work such as DeiT~\cite{touvron2021deit} and Swin~\cite{liu2021swin} showed that training protocol and hierarchical design strongly affect downstream performance. Our work does not propose a new backbone; instead, it modifies the FFN inside a fixed ViT-style block.

\paragraph{Expressive FFN replacements.}
Gated activations and multiplicative channel mixers, such as GLU variants~\cite{shazeer2020glu}, increase expressivity by introducing feature-wise products. Recent vision-specific bilinear operators go further and directly model second-order interactions. The most relevant baseline in our study is \AFBO~\cite{afbo2025}, which replaces the FFN with an asymmetric factorized bilinear operator built from grouped channel mappings.

\paragraph{Residual readout and depth aggregation.}
We also use a penultimate-residual (\PR) readout, which forms the final classifier representation as a residual combination of the last and penultimate layer states. In this paper that readout is treated as a fixed control shared by all \AFBO-based variants; it is not the main contribution. Our goal is to understand how much additional signal remains after the dominant final feature is fixed, and whether orthogonal quadratic complements can exploit it more effectively than unrestricted auxiliary branches.

\section{Method}
\subsection{Theoretical Motivation: Beyond Linear Subspaces}

\paragraph{The need for explicit feature interaction.}
Standard vision transformer FFNs operate as token-wise multi-layer perceptrons (MLPs). Their primary limitation is the lack of explicit high-order feature interaction; they must rely on compositional depth to indirectly model complex dependencies. Directly injecting a quadratic auxiliary branch introduces structural curvature and multiplicative interactions, akin to the mechanisms that make Gated Linear Units (GLU) and bilinear pooling effective. From a functional perspective, a standard fully-connected layer learns mappings confined to a linear subspace, $\mathrm{span}(W)$. A quadratic auxiliary branch, conversely, models interactions of the form $x^\top A x$. These represent fundamentally distinct function families. The quadratic component is specifically designed to capture structural information strictly outside the main linear subspace.

\paragraph{The necessity of redundancy removal.}
However, simply adding an unrestricted high-order branch does not guarantee that the network will learn non-redundant information. Consider a typical formulation of a quadratic feature such as $q(x) = (U x) \odot (V x)$. When expanded, a portion of this product captures genuine cross-feature interactions, but another substantial component often reduces to a linear deformation proportional to a standard mapping $W x$.

Suppose a transformer block already possesses a dominant hidden map $b(x) \in \mathbb{R}^{C \times H \times W}$ produced by a strong bilinear FFN such as \AFBO. A naive additive auxiliary branch would inject $q(x)$ as
\begin{equation}
  y = b(x) + \alpha\, q(x).
\end{equation}
If the redundancy is not explicitly managed, the network frequently collapses to a lazy solution where the auxiliary branch mimics the dominant map, yielding $y \approx (1 + \alpha') b(x)$. The auxiliary branch fails to introduce new structural information and merely amplifies the existing host representation.

To solve this, we interpret $b(x)$ and the auxiliary feature as elements of a finite-dimensional Hilbert space equipped with the standard inner product. By explicitly defining the main branch as the dominant subspace, we can project the auxiliary feature onto the orthogonal complement of that space. Cutting out the linearly dependent duplicate leaves exactly the genuine, non-redundant higher-order variations.

\subsection{Main Branch Representation}
Our method is designed to be agnostic to the specific architecture of the host FFN. Let the host FFN produce a dominant hidden representation $b(x)$. The \OQC\ module can be integrated into any host: standard MLPs, bilinear operators such as \AFBO~\cite{afbo2025}, or other FFN variants. In our primary experiments we use \AFBO\ as the host because it already provides a strong dominant representation, making it a demanding testbed for the complement; we also evaluate directly on a plain MLP host in Section~\ref{sec:generic}. In our implementation, this hidden map is then fed to an output projection and wrapped inside a standard pre-norm transformer block. We keep that host branch fixed and only change the auxiliary pathway. This isolates the contribution of the complement rather than conflating it with a different backbone.

\subsection{Orthogonal Quadratic Complements}
We first construct a low-rank quadratic auxiliary feature. Let
\begin{equation}
  u(x), v(x) \in \mathbb{R}^{r \times H \times W}
\end{equation}
be two learned projections of the block input, and define the raw quadratic feature
\begin{equation}
  q(x) = \mathrm{RMSNorm}\!\left(u(x) \odot v(x)\right),
\end{equation}
where $\odot$ denotes Hadamard multiplication and $r \ll C$.

To orthogonalize this feature with respect to the main branch, we map the main hidden map into the same auxiliary space:
\begin{equation}
  m(x) = \mathrm{RMSNorm}\!\left(P\, b(x)\right).
\end{equation}
Here $P: \mathbb{R}^{C \times H \times W} \to \mathbb{R}^{r \times H \times W}$ is a learned projection into the auxiliary space.
We then remove the component aligned with $m(x)$:
\begin{equation}
  q^\perp(x)
  =
  \mathrm{RMSNorm}\!\left(
    q(x)
    -
    \frac{\langle q(x), m(x)\rangle}{\lVert m(x)\rVert^2 + \varepsilon}
    m(x)
  \right).
  \label{eq:oqc}
\end{equation}

\paragraph{Orthogonality property.}
For every nonzero $m(x)$, the unnormalized residual in Eq.~\eqref{eq:oqc} is orthogonal to $m(x)$ by construction:
\begin{equation}
  \left\langle
  q(x) - \frac{\langle q(x), m(x)\rangle}{\lVert m(x)\rVert^2 + \varepsilon} m(x),
  \, m(x)
  \right\rangle
  \approx 0.
\end{equation}
Our mechanism study confirms that the absolute cosine overlap between auxiliary and main features drops from roughly $0.13$--$0.14$ before projection to approximately $10^{-8}$ after projection.

\paragraph{Interpretation.}
This projection is the core distinction between our method and a naive auxiliary quadratic branch. A standard additive auxiliary branch is free to relearn directions already present in the host representation. In contrast, \OQC\ forces the auxiliary branch to act as a complement. In Hilbert-space terms, the main branch defines the dominant subspace and the quadratic branch is restricted to the corresponding orthogonal complement. This gives the method a precise non-redundancy bias rather than a generic ``more capacity'' story.

\subsection{Full \OQC\ and Efficient \OQCLR}
We study two realizations of the complement.

\paragraph{Full \OQC.}
The first lifts the auxiliary feature to the full hidden width, performs the complement in hidden space, and adds the resulting update to the main branch:
\begin{equation}
  h(x) = b(x) + \sigma(\beta)\, O\!\left(q^\perp(x)\right),
\end{equation}
where $O$ maps the auxiliary feature to the hidden width and $\sigma(\beta)$ is a learned mixing coefficient.

\paragraph{\OQCLR.}
The second performs orthogonalization entirely in low-rank space and only then lifts the result back to the hidden width:
\begin{equation}
  h(x) = b(x) + \sigma(\beta)\, \Delta(x),
\end{equation}
with
\begin{equation}
  \Delta(x)=\mathrm{RMSNorm}\!\left(O(q^\perp(x))\right).
\end{equation}
This \OQCLR\ form preserves most of the gain while substantially reducing the runtime gap relative to full \OQC. Empirically, the best trade-off in our current sweep is rank $56$, which we denote as \OQCLR\ r56.

\subsection{Gated \OQC\ Variants}
We also examine gated complement variants, which modulate the orthogonal auxiliary update after projection.

\paragraph{\OQC-static.}
A static scalar gate uses the same per-model learned mixing coefficient $\sigma(\beta)$ as \OQCLR, but with a more conservative initialization ($\beta_0 \approx 0$) and an explicit gate parameterization that is decoupled from the host FFN design:
\begin{equation}
  h(x) = b(x) + \sigma(\beta)\, \Delta(x), \quad \beta_0 \sim \mathcal{N}(0, 0.01).
\end{equation}
The key distinction from \OQCLR\ is architectural independence: the static gate is implemented as a standalone parameter rather than being embedded in the host FFN class, allowing the same gate design to be reused across arbitrary host branches.

\paragraph{\OQC-dynamic.}
A dynamic gate predicts an input-dependent spatial modulation map $g(x)$ via a lightweight $1\times1$ convolution on the input:
\begin{equation}
  h(x) = b(x) + \sigma(g(x)) \odot \Delta(x).
\end{equation}
This increases adaptivity by allowing the complement injection to vary per token and per sample. In our experiments, the dynamic gate achieves competitive accuracy ($65.33 \pm 0.11\%$, single protocol) while maintaining a low mean activation ($\sim 0.11$) with a non-trivial standard deviation ($0.121$), indicating genuinely input-dependent modulation rather than a reparameterized constant.

\subsection{Architecture Illustration}
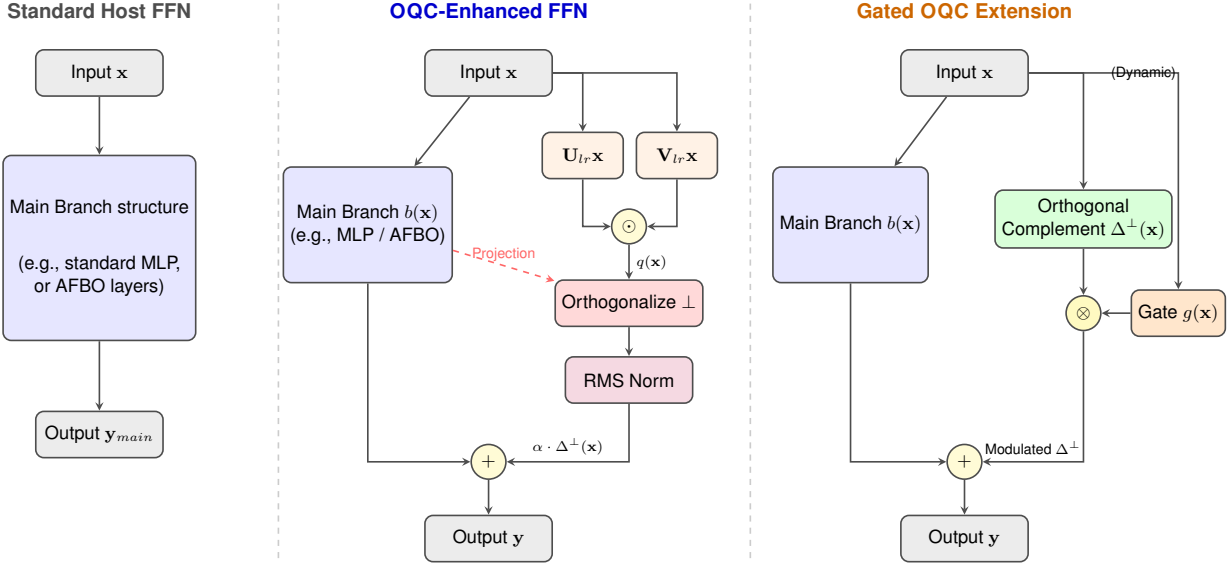
\begin{figure*}[htbp]
\centering
\resizebox{0.92\textwidth}{!}{%
\begin{tikzpicture}[
    node distance=0.6cm and 0.8cm,
    box/.style={rectangle, rounded corners, draw=black!70, thick, minimum width=2.2cm, minimum height=0.8cm, align=center, font=\small\sffamily},
    mainbox/.style={box, fill=blue!10},
    oqcbox/.style={box, fill=orange!10},
    opbox/.style={circle, draw=black!70, thick, fill=yellow!20, minimum size=0.6cm, inner sep=0pt, font=\small\sffamily},
    arrow/.style={->, >=stealth, thick, draw=black!70}
]

\node (x_std) [box, fill=gray!15] {Input $\mathbf{x}$};
\node (main_std) [mainbox, below=1.0cm of x_std, minimum height=3.2cm] {Main Branch structure\\ \vspace{0.2cm} \\(e.g., standard MLP,\\or AFBO layers)};
\node (y_std) [box, fill=gray!15, below=1.2cm of main_std] {Output $\mathbf{y}_{main}$};

\draw [arrow] (x_std) -- (main_std);
\draw [arrow] (main_std) -- (y_std);

\node[above=0.4cm of x_std, font=\bfseries\sffamily, text=black!70] {Standard Host FFN};

\node (x_oqc) [box, fill=gray!15, right=4.5cm of x_std] {Input $\mathbf{x}$};

\node (main_oqc) [mainbox, below left=1.2cm and -0.5cm of x_oqc, minimum height=2.0cm] {Main Branch $b(\mathbf{x})$\\(e.g., MLP / AFBO)};

\node (u_oqc) [oqcbox, minimum width=1.4cm, below right=0.6cm and -0.2cm of x_oqc] {$\mathbf{U}_{lr}\mathbf{x}$};
\node (v_oqc) [oqcbox, minimum width=1.4cm, right=0.2cm of u_oqc] {$\mathbf{V}_{lr}\mathbf{x}$};
\node (hadamard_oqc) [opbox, below=0.5cm of u_oqc, xshift=0.8cm] {$\odot$};
\node (ortho_oqc) [oqcbox, fill=red!15, below=0.6cm of hadamard_oqc] {Orthogonalize $\perp$};
\node (rms_oqc) [oqcbox, fill=purple!15, below=0.5cm of ortho_oqc] {RMS Norm};

\node (add_oqc) [opbox, below=6.0cm of x_oqc] {$\mathbf{+}$};
\node (y_oqc) [box, fill=gray!15, below=0.6cm of add_oqc] {Output $\mathbf{y}$};

\draw [arrow] (x_oqc) -- (main_oqc);
\draw [arrow] (x_oqc) -| (u_oqc);
\draw [arrow] (x_oqc) -| (v_oqc);
\draw [arrow] (u_oqc) |- (hadamard_oqc);
\draw [arrow] (v_oqc) |- (hadamard_oqc);
\draw [arrow] (hadamard_oqc) -- (ortho_oqc) node[midway, right, font=\scriptsize\sffamily] {$q(\mathbf{x})$};
\draw [arrow] (ortho_oqc) -- (rms_oqc);
\draw [arrow] (main_oqc) |- (add_oqc);
\draw [arrow] (rms_oqc) |- (add_oqc) node[near end, above, font=\scriptsize\sffamily] {$\alpha \cdot \Delta^\perp(\mathbf{x})$};
\draw [arrow] (add_oqc) -- (y_oqc);

\draw [arrow, dashed, draw=red!60, thick] (main_oqc) -- (ortho_oqc) node[midway, above=-0.05cm, font=\scriptsize\sffamily, text=red!60] {Projection};

\node[above=0.4cm of x_oqc, font=\bfseries\sffamily, text=blue!80!black] {\OQC{}-Enhanced FFN};

\node (x_gated) [box, fill=gray!15, right=6.0cm of x_oqc] {Input $\mathbf{x}$};

\node (main_gated) [mainbox, below left=1.2cm and -0.5cm of x_gated, minimum height=2.0cm] {Main Branch $b(\mathbf{x})$};

\node (oqc_gated) [oqcbox, fill=green!15, below right=1.6cm and -0.6cm of x_gated, minimum height=1.0cm] {Orthogonal\\Complement $\Delta^\perp(\mathbf{x})$};

\node (gate_op) [opbox, fill=yellow!30, below=0.8cm of oqc_gated] {$\otimes$};
\node (gate_net) [box, fill=orange!20, right=0.5cm of gate_op, minimum width=1.4cm] {Gate $g(\mathbf{x})$};

\node (add_gated) [opbox, below=6.0cm of x_gated] {$\mathbf{+}$};
\node (y_gated) [box, fill=gray!15, below=0.6cm of add_gated] {Output $\mathbf{y}$};

\draw [arrow] (x_gated) -- (main_gated);
\draw [arrow] (x_gated) -| (oqc_gated);
\draw [arrow] (x_gated) -| (gate_net) node[near start, right, font=\scriptsize\sffamily] {(Dynamic)};

\draw [arrow] (oqc_gated) -- (gate_op);
\draw [arrow] (gate_net) -- (gate_op);
\draw [arrow] (main_gated) |- (add_gated);
\draw [arrow] (gate_op) |- (add_gated) node[near end, above, font=\scriptsize\sffamily] {Modulated $\Delta^\perp$};
\draw [arrow] (add_gated) -- (y_gated);

\node[above=0.4cm of x_gated, font=\bfseries\sffamily, text=orange!80!black] {Gated \OQC{} Extension};

\draw [dashed, draw=gray!40, thick] ($(x_std)!0.46!(x_oqc)$) +(0, 1.2) -- +(0, -8.5);
\draw [dashed, draw=gray!40, thick] ($(x_oqc)!0.55!(x_gated)$) +(0, 1.2) -- +(0, -8.5);

\end{tikzpicture}
}
\caption{Architecture evolution. \textbf{Left:} The host FFN (e.g., standard MLP or \AFBO) produces a dominant main branch. \textbf{Middle:} \OQC\ adds a low-rank quadratic auxiliary feature and explicitly projects it onto the orthogonal complement of the main branch. \textbf{Right:} The gated \OQC\ family maintains the orthogonal complement but dynamically or statically modulates its injection to the host branch.}
\label{fig:oqc-arch}
\end{figure*}

Figure~\ref{fig:oqc-arch} summarizes the progression from a standard host FFN to \OQC\ and finally to the gated \OQC\ extension. The visual difference reveals the core mechanism: the host FFN (e.g., an MLP or \AFBO) produces a dominant main branch, while \OQC\ introduces a low-rank quadratic auxiliary feature and explicitly projects it onto the orthogonal complement of the host's representation. The gated \OQC\ family preserves this orthogonal complement but conditionally modulates it using a static or dynamic gate.

\subsection{Penultimate-Residual Readout}
All main \AFBO-based experiments use the same readout:
\begin{equation}
  z = h_L + \sigma(\gamma) h_{L-1},
\end{equation}
where $h_L$ and $h_{L-1}$ are the last and penultimate layer representations. We keep this component fixed across the \AFBO, full \OQC, and \OQCg\ comparisons so that the paper focuses on the FFN change itself rather than on readout engineering.

\section{Experiments}
\subsection{Experimental Protocol}
Our primary protocol uses Deep-ViT on CIFAR-100 with depth $8$, width $256$, $8$ heads, patch size $4$, batch size $512$, and three seeds. We train with AdamW, a peak learning rate of $2\times10^{-3}$, weight decay $0.05$, and a short warmup before cosine decay. We report the best test accuracy over training, parameter count, and measured throughput in images per second. The TinyImageNet transfer uses the same backbone family with image size $64$, patch size $8$, batch size $128$, and three seeds. Unless otherwise noted, all \AFBO-based methods share the same penultimate-residual readout, so differences in Tables~\ref{tab:cifar-main}--\ref{tab:generic} isolate the FFN change itself.

\subsection{Main Results on Deep-ViT/CIFAR-100}
\begin{table}[t]
  \centering
  \small
  \begin{tabular}{lccc}
    \toprule
    Method & Acc. (\%) & Params (M) & Img/s \\
    \midrule
    \AFBO+\PR & 64.25 $\pm$ 0.22 & 6.37 & \textbf{9762} \\
    \OQC\ full & \textbf{65.59 $\pm$ 0.22} & 6.97 & 5642 \\
    \OQCLR\ r56 & 65.52 $\pm$ 0.25 & 7.19 & 7497 \\
    \OQC-static & 65.44 $\pm$ 0.19 & 7.59 & 7032 \\
    \OQC-dynamic & 65.33 $\pm$ 0.11 & 7.59 & 7155 \\
    \bottomrule
  \end{tabular}
  \caption{Main CIFAR-100 results under the common Deep-ViT protocol. Full \OQC\ is the strongest cross-validated variant, while \OQCLR\ r56 offers a better speed--accuracy trade-off.}
  \label{tab:cifar-main}
\end{table}

Table~\ref{tab:cifar-main} establishes the main empirical hierarchy. Full \OQC\ is the strongest overall variant and improves the matched \AFBO\ baseline by $+1.34$ points. \OQCLR\ r56 recovers nearly all of that gain while being markedly faster than full \OQC, making it the strongest efficiency-oriented version in the current suite. The gated \OQC\ family is competitive on CIFAR-100: \OQC-static\ slightly outperforms \OQC-dynamic\ in accuracy, while the dynamic gate remains somewhat faster and later shows the strongest class separation signal.

Two points are worth emphasizing. First, these gains are not coming from a readout trick alone, because the readout is fixed across all methods. Second, the variants represent different trade-offs rather than a single monotone frontier: full \OQC\ is the best geometry-preserving complement, \OQCLR\ r56 is the best speed--accuracy compromise, and \OQC-static\ is the strongest gated instance on CIFAR-100 under the main protocol. The TinyImageNet experiment (Section~\ref{sec:tiny}) later shows that \OQC-dynamic\ takes the lead on the second dataset.

\subsection{Cross-Dataset Transfer on TinyImageNet}
\label{sec:tiny}
\begin{table}[t]
  \centering
  \small
  \begin{tabular}{lccc}
    \toprule
    Method & Acc. (\%) & Params (M) & Img/s \\
    \midrule
    \AFBO+\PR          & 50.45 $\pm$ 0.21           & 6.43 & \textbf{11900} \\
    \OQC\ full         & 51.36 $\pm$ 0.28           & 7.03 & 8085 \\
    \OQCLR\ r56   & 51.35 $\pm$ 0.11           & 7.25 & 9400 \\
    \OQC-static        & 51.47 $\pm$ 0.07           & 7.65 & 9108 \\
    \OQC-dynamic       & \textbf{51.88 $\pm$ 0.32}  & 7.65 & 9051 \\
    \bottomrule
  \end{tabular}
  \caption{TinyImageNet cross-dataset transfer (3 seeds, consistent protocol). All \OQC\ variants improve the \AFBO\ baseline. \OQC-dynamic\ achieves the highest accuracy, confirming that the gated family generalizes beyond CIFAR-100.}
  \label{tab:tiny}
\end{table}

The TinyImageNet transfer in Table~\ref{tab:tiny} validates the \OQC\ family at scale and across datasets. All four complement variants consistently improve the \AFBO\ baseline, with gains ranging from $+0.90$ (\OQCLR\ r56) to $+1.43$ (\OQC-dynamic) percentage points. Notably, \OQC-dynamic\ achieves $51.88 \pm 0.32$---the highest single result on TinyImageNet---while \OQC-static\ achieves the tightest variance ($\pm 0.07$), indicating reliable improvement across seeds.

These results resolve the earlier uncertainty about cross-dataset generalization of the gated family. Both \OQC-static\ and \OQC-dynamic\ transfer positively to TinyImageNet, demonstrating that adaptive gating is not an artifact of CIFAR-100-specific tuning.

\subsection{Mechanism Analysis}
\begin{table}[t]
  \centering
  \small
  \begin{tabular}{lcccc}
    \toprule
    Method & Acc. & EffRank & PartRatio & Sep. \\
    \midrule
    \AFBO+\PR & 64.41 & 76.64 & 48.49 & 1.0118 \\
    \OQC\ full & 65.18 & \textbf{78.17} & \textbf{50.43} & 1.0146 \\
    \OQCLR\ r56 & 65.29 & 77.53 & 49.85 & 1.0194 \\
    \OQC-static & \textbf{65.44} & 77.67 & 50.05 & 1.0172 \\
    \OQC-dynamic & 65.33 & 77.49 & 49.96 & \textbf{1.0202} \\
    \bottomrule
  \end{tabular}
  \caption{Mechanism statistics on CIFAR-100. ``Sep.'' denotes class separation score.}
  \label{tab:mechanism}
\end{table}

Table~\ref{tab:mechanism} reveals a useful split between geometry and discrimination. Full \OQC\ preserves the richest representation geometry, with the best effective rank and participation ratio. The gated family slightly sacrifices that geometry but improves class separation, which explains why \OQC-static\ and \OQC-dynamic\ remain competitive in accuracy on CIFAR-100.

Orthogonalization itself behaves exactly as intended. For \OQCLR\ r56, the mean auxiliary--main absolute cosine drops from $0.143$ before projection to $1.49 \times 10^{-8}$ after projection. For \OQC-static\ and \OQC-dynamic, the corresponding drops are $0.133 \rightarrow 1.23 \times 10^{-8}$ and $0.131 \rightarrow 1.23 \times 10^{-8}$. The dynamic gate also changes how much complement is used: its mean mixing coefficient decreases from roughly $0.190$ to $0.130$, while the gate standard deviation rises to $0.121$, indicating genuinely input-dependent modulation rather than a disguised constant scalar.

\begin{figure}[!ht]
\centering
\includegraphics[width=\linewidth]{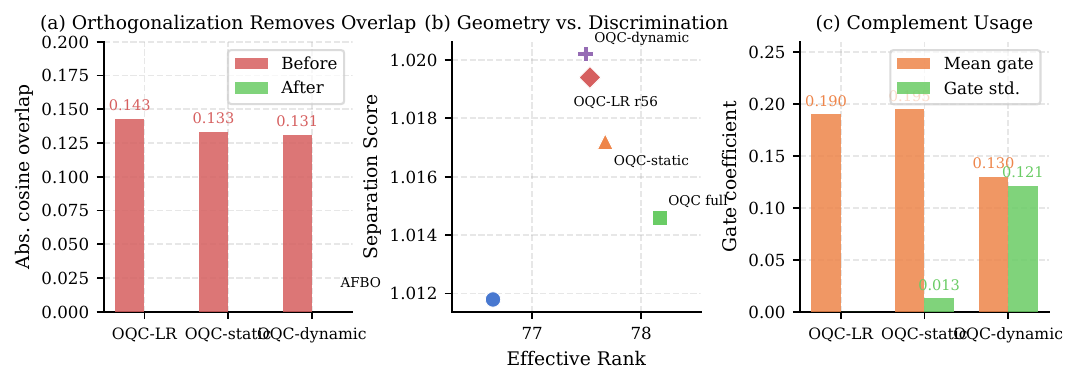}
\caption{Mechanism visualization. \textbf{(a)} Auxiliary--main absolute cosine overlap before and after complement projection; orthogonalization reduces it to machine-precision zero. \textbf{(b)} Effective-rank vs.\ separation score: full \OQC\ achieves the richest geometry while \OQC-dynamic\ achieves the best class separation. \textbf{(c)} Complement usage: \OQC-dynamic\ lowers average gate activation and introduces genuine input-dependent variation (high gate std.), confirming it is not a reparameterized static gate.}
\label{fig:oqc-mechanism}
\end{figure}

The mechanism picture therefore matches the accuracy picture. Full \OQC\ is the most geometry-preserving variant, while \OQC-dynamic\ is the most discriminative according to separation score. \OQC-static\ ends up sitting between those extremes and is therefore the strongest gated accuracy point in this specific protocol. Figure~\ref{fig:oqc-mechanism} makes this split more visible: panel~(a) shows that overlap is almost annihilated after projection, panel~(b) shows that full \OQC\ occupies the best geometry corner while \OQC-dynamic\ pushes furthest toward discrimination, and panel~(c) shows that dynamic gating genuinely changes complement usage rather than merely reparameterizing the same scalar gate.

\subsection{Ablation on Auxiliary Rank and Design}
To understand the sensitivity of \OQC\ to its capacity and architectural choices, we conducted several ablation studies on CIFAR-100.

\begin{figure}[!ht]
\centering
\includegraphics[width=\linewidth]{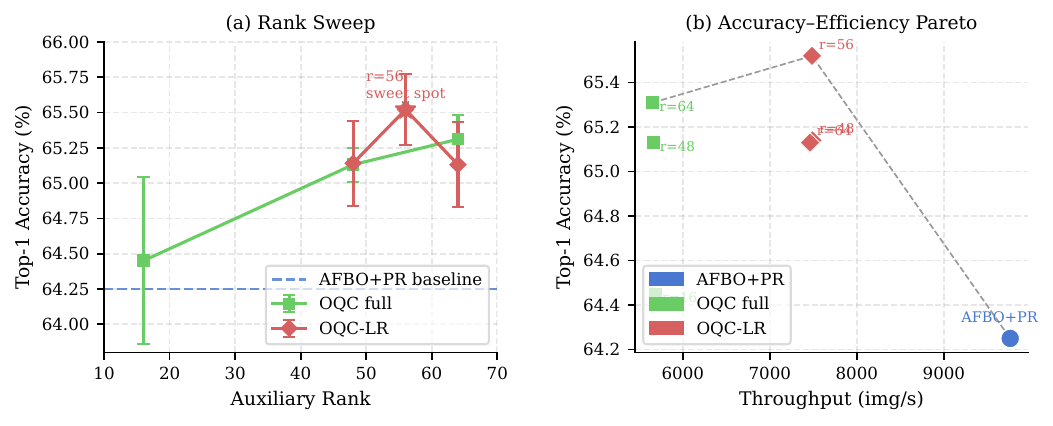}
\caption{Accuracy--efficiency trade-off. \textbf{(a)} Rank sweep: full \OQC\ improves monotonically with rank, while \OQCLR\ peaks at $r{=}56$ (starred). \textbf{(b)} Pareto frontier: \OQCLR\ r56 delivers $65.52\%$ at $7{,}485$~img/s, substantially faster than full \OQC\ ($5{,}655$~img/s) while nearly matching its accuracy.}
\label{fig:oqc-rank}
\end{figure}

\paragraph{Auxiliary Rank Sweep.}
We varied the rank $r$ of the auxiliary quadratic branch.
Starting from a baseline accuracy of $64.25\%$, an extremely low rank ($r=16$) yields $64.45 \pm 0.59\%$---only a marginal improvement---while rank $r=48$ and $r=64$ climb to $65.13 \pm 0.12\%$ and $65.31 \pm 0.17\%$ respectively.
This confirms that the orthogonal complement needs sufficient dimensionality to capture non-redundant high-order features effectively.
Figure~\ref{fig:oqc-rank} shows the resulting trade-off: the full complement improves steadily with rank (left panel), while \OQCLR\ peaks at $r=56$ and forms the cleanest efficiency sweet spot in the Pareto view (right panel).

\paragraph{Alternative Complement Designs.}
We explored alternative, cheaper designs for the auxiliary branch to validate the necessity of each component in \OQC:
\begin{itemize}[leftmargin=*,itemsep=0pt,topsep=2pt]
\item \textbf{Shared Projection}: Sharing the projection matrices between the main and auxiliary branches reduces the parameter overhead but drops the accuracy to $64.85 \pm 0.14\%$, indicating that the auxiliary branch needs independent capacity to find complementary features.
\item \textbf{Scalar Gated (No Orthogonalization)}: A naive scalar-gated auxiliary branch without explicit orthogonalization only achieves $63.47 \pm 0.19\%$, performing worse than the baseline. This strongly validates our core hypothesis: simply adding more quadratic capacity is ineffective; the \emph{orthogonalization} step is what prevents redundancy and drives the performance gain.
\item \textbf{Low-Rank Orthogonalization (No Gate)}: When orthogonalization is performed in low-rank space but the result is injected without any gating ($\alpha \equiv 1$), accuracy drops to $63.03 \pm 0.23\%$ (3-seed confirmation run). This demonstrates that while orthogonalization provides non-redundant features, a gating mechanism is strictly necessary to control the injection scale and prevent the complement from destabilizing the main branch.
\end{itemize}

\noindent
\emph{Note.} The main results in Table~\ref{tab:cifar-main} and the ablation confirmation above use the same Deep-ViT backbone but slightly different training protocols (learning rate and total epoch budget). Numbers within each protocol are directly comparable; cross-protocol comparisons should account for this difference.

\subsection{Generic Transfer and Component Decomposition}
\label{sec:generic}
\begin{table}[t]
  \centering
  \small
  \setlength{\tabcolsep}{6pt}
  \begin{tabular}{lcc}
    \toprule
    Condition & MLP & \AFBO \\
    \midrule
    Base                & 47.27\,$\pm$\,0.42 & 62.82\,$\pm$\,0.45 \\
    $+$\PR              & 47.29\,$\pm$\,0.25 & 63.77\,$\pm$\,0.11 \\
    $+$\OQCLR           & 50.75\,$\pm$\,0.23 & 63.88\,$\pm$\,0.29 \\
    $+$\OQCLR$+$\PR     & 51.39\,$\pm$\,0.42 & \textbf{65.16\,$\pm$\,0.07} \\
    \bottomrule
  \end{tabular}
  \caption{Component decomposition on CIFAR-100 (3 seeds, top-1 acc.\%).
    \emph{Base}: host FFN only, no readout, no complement.
    \PR: penultimate-residual readout.
    \OQCLR: efficient low-rank complement.}
  \label{tab:generic}
\end{table}

Table~\ref{tab:generic} decomposes the two design choices---complement (\OQC) and readout (\PR)---across two host architectures. Three findings emerge.

\textbf{PR readout is structure-dependent.}
The penultimate-residual readout adds negligible gain for a plain MLP ($+0.02$ pp) but a meaningful $+0.95$ pp for the richer \AFBO\ branch, confirming that the readout exploits depth-aggregated structure that a simple MLP does not produce.

\textbf{OQC-LR is broadly applicable.}
The complement improves both hosts: MLP gains $+3.48$ pp and \AFBO\ gains $+1.06$ pp without any readout. Gains are inversely proportional to host strength, consistent with the intuition that \OQC\ is most useful when the host defines a stable dominant subspace while still leaving orthogonal second-order directions unexploited.

\textbf{OQC and PR are super-additive on \AFBO.}
Combined, \OQCLR$+$\PR adds $+2.34$ pp from the plain \AFBO\ baseline, exceeding the na\"{i}ve sum of their individual contributions ($0.95{+}1.06{=}2.01$ pp). We hypothesize that the complement enriches the depth-token distribution in a way that the PR readout can exploit more effectively.

\subsection{Backbone Transfer Caveat}
We also ran a small ViT-Tiny transfer study for the penultimate-residual readout in isolation. The result was mixed rather than clearly positive. We do not include that setting as a main table because it does not test the full \OQC\ method, but it is still informative: simple readout improvements should not be confused with the complement mechanism itself. This is another reason the paper focuses on the \AFBO-hosted complement family rather than on a universal readout claim.

\section{Discussion and Limitations}
Three limitations matter for interpretation.

\paragraph{The complement is not universal.}
Our generic transfer experiment shows a large positive result on a plain MLP host and a moderate gain on \AFBO. We do not claim that orthogonal quadratic complements should be inserted into every FFN indiscriminately; in particular, the gain diminishes for hosts that already model strong quadratic interactions.

\paragraph{Gated variants generalize, but with wider variance.}
Our TinyImageNet transfer experiment confirms that both \OQC-static\ and \OQC-dynamic\ produce consistent gains on a second dataset. However, \OQC-dynamic\ shows a higher standard deviation ($\pm 0.32$) compared to \OQC-static\ ($\pm 0.07$) and the ungated variants. This suggests that dynamic gating may be sensitive to initialization, and that additional runs or learning-rate tuning could be worthwhile before treating \OQC-dynamic\ as the definitive leader.

\paragraph{Readout gains do not appear universally backbone-agnostic.}
In a separate ViT-Tiny pilot, penultimate-residual readout alone did not improve the baseline ($24.37 \pm 0.95$ vs.\ $23.94 \pm 0.38$). For this reason, we use the readout as a controlled component inside the \AFBO\ family, but avoid promoting it as a universal architectural improvement.

\section{Conclusion}
We introduced \OQC, a feed-forward design principle that injects only the orthogonal quadratic complement of a dominant main branch. The resulting method improves a matched \AFBO\ baseline on Deep-ViT/CIFAR-100 and TinyImageNet across all tested variants---ungated and gated alike. Full \OQC\ and \OQCLR\ provide the most stable improvements; \OQC-dynamic\ achieves the highest single accuracy on TinyImageNet ($51.88 \pm 0.32$), and mechanism analyses confirm that the auxiliary branch becomes non-redundant after projection. \OQCLR\ is the most practical efficiency-oriented variant; and the \OQCg\ family is a cross-dataset-validated extension that trades some geometric richness for stronger class discrimination. More broadly, the results suggest that second-order auxiliary branches are most useful when they are explicitly constrained to occupy the complement of the dominant representation rather than to compete with it directly.

\end{document}